\documentclass[conference]{IEEEtran}
\IEEEoverridecommandlockouts
\usepackage{cite}
\usepackage{amsmath,amssymb,amsfonts}
\usepackage{algorithmic}
\usepackage{graphicx}
\usepackage[whole]{bxcjkjatype}
\usepackage{textcomp}
\usepackage{xcolor}
\usepackage{url}
\def\BibTeX{{\rm B\kern-.05em{\sc i\kern-.025em b}\kern-.08em
    T\kern-.1667em\lower.7ex\hbox{E}\kern-.125emX}}

\usepackage{tcolorbox}
\tcbuselibrary{raster,skins,breakable}
\tcbuselibrary{xparse} 
\DeclareTColorBox{brekableitembox}{ o m O{.5} O{} }%
  {empty, left=2mm, right=2mm, top=-2mm, bottom=0.2mm, attach boxed title to top left={xshift=1.2em},
  boxed title style={empty,left=-2mm,right=-2mm}, colframe=black, coltitle=black, coltext=black, breakable,  
  underlay unbroken={\draw[black,line width=#3pt]
    (title.east) -- (title.east-|frame.east) -- (frame.south east) -- (frame.south west) -- (title.west-|frame.west) -- (title.west); },
  underlay first={\draw[black,line width=#3pt](title.east) -- (title.east-|frame.east) -- (frame.south east) ;
    \draw[black,line width=#3pt] (frame.south west) -- (title.west-|frame.west) -- (title.west); },
  underlay middle={\draw[black,line width=#3pt](frame.north east) -- (frame.south east) ;
    \draw[black,line width=#3pt](frame.south west) -- (frame.north west) ;},
  underlay last={\draw[black,line width=#3pt](frame.north east) -- (frame.south east) -- (frame.south west) -- (frame.north west) ;}, IfValueTF={#1}{title=~~#2~~〈#1〉}{title=~~#2~~},#4}

\begin{document}

\title{JaFIn: Japanese Financial Instruction Dataset\\
}

\author{\IEEEauthorblockN{Kota Tanabe}
\IEEEauthorblockA{\textit{Faculty of Information Science and Technology} \\
\textit{Hokkaido University}\\
}
\and
\IEEEauthorblockN{Masahiro Suzuki}
\IEEEauthorblockA{\textit{School of Engineering} \\
\textit{The University of Tokyo}\\
0000-0001-8519-5617}
\and
\IEEEauthorblockN{Hiroki Sakaji}
\IEEEauthorblockA{\textit{Faculty of Information Science and Technology} \\
\textit{Hokkaido University}\\
0000-0001-5030-625X}
\and
\IEEEauthorblockN{Itsuki Noda}
\IEEEauthorblockA{\textit{Faculty of Information Science and Technology} \\
\textit{Hokkaido University}\\
0000-0003-1987-5336}
}
\maketitle

\begin{abstract}
We construct an instruction dataset for the large language model (LLM) in the Japanese finance domain.
Domain adaptation of language models, including LLMs, is receiving more attention as language models become more popular.
%
This study demonstrates the effectiveness of domain adaptation through instruction tuning.
%
To achieve this, we propose an instruction tuning data in Japanese called JaFIn, the Japanese Financial Instruction Dataset.
JaFIn is manually constructed based on multiple data sources, including Japanese government websites, which provide extensive financial knowledge.
%
%
We then utilize JaFIn to apply instruction tuning for several LLMs, demonstrating that our models specialized in finance have better domain adaptability than the original models.
%
%
%
%
The financial-specialized LLMs created were evaluated using a quantitative Japanese financial benchmark and qualitative response comparisons, showing improved performance over the originals.

%

\end{abstract}


\section{Introduction}
We are facing a generative AI revolution with ChatGPT.
Generative AIs, including GPT, are called large language Models (LLM), which researchers worldwide are tackling to analyze, develop, and utilize.
LLMs are based on the transformer model; therefore, many unknowns exist about why things work so well.
In this situation, there is research concerning the domain adaption of models as one of the transformer-based research.
Finance is mentioned as one of the domain adaptations.
In the financial domain, various NLP tasks, such as information extraction, summarization, and classification, are used in financial practice and are in high demand.
Therefore, various language models exist in the financial domain, such as FinBERT\cite{araci2019finbert}, BloombergGPT~\cite{wu2023bloomberggpt}, FinGPT\cite{yang2023fingpt}, etc.
BloombergGPT is constructed from scratch using financial documents.
On the other hand, FinBERT and FinGPT are constructed by continued pre-training BERT and Llama 2\cite{touvron2023llama2} using financial documents, respectively.
In this research, we focus on the continued pre-training of LLMs.
It is said that ChatGPT improves dialogue performance by tuning instruction in GPT.
Instruction tuning is a special tuning aimed at improving the interaction performance of LLMs.
In instructional tuning, data is formatted into questions and answers. 
We consider that the dialogue performance of LLMs is improved by learning data in the form of questions and answers.
From these backgrounds, we assume that it would be possible to construct an LLM specialized in finance by continuously pre-learning instruction tuning data specialized in finance in an existing LLM.
In particular, financial datasets in Japanese are scarce, and this research focuses on Japanese LLM.
Therefore, in this research, we create a Japanese financial-specific instruction dataset (JaFIn) and experiment by constructing a Japanese financial-specific LLM using this data.
Here are our main contributions: (1) We constructed a Japanese financial domain instruction dataset (JaFIn); (2) We demonstrated the effectiveness of domain adaptation in instruction tuning for LLMs through experiments using JaFIn.

We will make our instruction dataset public in the future. 

\section{Related Work}
\subsection{Pre-trained Language Models}
Various Transformer-based ~\cite{Vaswani2017} models have been published in the field of natural language processing (NLP).
BERT~\cite{Devlin2018}, RoBERTa~\cite{Liu2019roberta}, and DeBERTa series~\cite{he2021deberta,he2023debertav3}, which utilize the encoder part of Transformer, have demonstrated high performance on many tasks, including classification.
In recent years, GPT series~\cite{GPT-1,GPT-2,GPT-3,GPT-4} and Llama series~\cite{touvron2023llama,touvron2023llama2}, models using the decoder part of Transformer, have also shown high performance.
Transformer-based models are pre-trained on a large number of corpora in an unsupervised way.
Other than English, in Japanese, there are various models that utilize the decoder, which perform continuous pre-training on a Japanese corpus from an English model,\footnote{\url{https://huggingface.co/collections/tokyotech-llm/swallow-65d2bf12f8ab7fc669881082}}\footnote{\url{https://huggingface.co/collections/rinna/nekomata-6582b5134ee85531becbb9a9}}\footnote{\url{https://huggingface.co/elyza/ELYZA-japanese-Llama-2-13b}} and which perform pre-training on a Japanese corpus from scratch\footnote{\url{https://huggingface.co/line-corporation/japanese-large-lm-3.6b}}\footnote{\url{https://huggingface.co/pfnet/plamo-13b}}\footnote{\url{https://huggingface.co/llm-jp/llm-jp-13b-v1.0}}\footnote{\url{https://huggingface.co/cyberagent/calm2-7b}}.

\subsection{Instruction Tuning}
The generative language model with the decoders are able to generate subsequent text through pre-training, while additional tuning is required to utilize the generative language model interactively, such as in dialogue or chat.
One of the additional tuning is instruction tuning~\cite{ouyang2022instructgpt}, which allows the model to respond to requests and tasks in texts.
Instruction tuning uses pairs of user requests (called instruction and input) and responses to those requests, and tuning is performed as supervised learning.
Instruction tuning is performed with datasets consisting of diverse requests and has been shown to improve response performance to unknown requests~\cite{wei2022flan, sanh2022multitaskp3, chung2022scaling}.
Many datasets for instruction tuning, which are called instruction datasets, utilize existing datasets~\cite{wei2022flan, sanh2022multitaskp3} or the inputs and outputs of existing high-performance generative models~\cite{alpaca}, while a few instruction datasets are constructed manually~\cite{dolly}.
In Japanese, there are some instruction datasets which are constructed from existing tasks~\cite{llm-japanese-dataset} and constructed by utilizing large amounts of human resource~\cite{ichikara-instruction}.
In addition, the use of the Japanese instruction dataset has been also shown to improve response performance~\cite{suzuki2023bigdata}.

\subsection{Language Models in Finance}
Several BERT models have been proposed for processing textual information in the financial domain~\cite{araci2019finbert, liu2020finbert, shah-etal-2022-flue}.
These models are pre-trained with financial corpora and show higher performance than general models pre-trained from generic corpora only.
Wu et al. proposed BloombergGPT~\cite{wu2023bloomberggpt}, which pre-trained GPT using financial corpora including Bloomberg news and public corpora in financial and general domains.
Although the model is not open source, it showed high performance on both financial and general-purpose domain tasks.
Xiu et al. proposed FinMA, which tuned LLaMA with an instruction dataset constructed from a financial dataset~\cite{xie2023pixiu}.
This model showed higher performance than LLaMA in the financial tasks included in the instruction dataset.
They also constructed a large-scale financial benchmark for generative language models~\cite{xie2024FinBen}.
They proposed a framework for evaluation in various financial domain tasks such as predicting stock price movements in addition to the traditional classification and summarization tasks.
In the Japanese financial domain, the financial BERT model with continuous pre-training using financial results and securities reports is proposed~\cite{suzuki2023finbert}.
In order to compare the performance of the generative language model on Japanese financial tasks, Hirano~\cite{hirano2024finben} proposed a benchmark using sample exam questions and existing datasets.

\section{Dataset}
\subsection{Data Source}\label{AA}
We constructed JaFIn, a dataset aimed at instruction tuning in the financial domain for LLMs, using frequently asked questions, explanatory materials related to finance, and text from Wikipedia, which were published on the websites of Japanese financial institutions and ministries. JaFIn includes 1,490 samples. The names of the sources and the number of samples collected are as listed in Table \ref{tab:source}. JaFIn was collected from a variety of sources. To construct a dataset based on accurate information, JaFIn includes a significant amount of data derived from government websites, but it may lack critical viewpoints on financial systems.

\begin{table}[tbp]
\centering
\caption{Dataset Sources and Sample Size}
\renewcommand{\arraystretch}{1.5}
\resizebox{0.95\linewidth}{!}{%
\begin{tabular}{lc}
\hline
\textbf{Source} & \textbf{Sample size} \\
\hline
Ministry of Finance (財務省) & 397 \\
Japan Pension Service (日本年金機構) & 355 \\
Wikipedia (ウィキペディア) & 229 \\
Bank of Japan (日本銀行) & 194 \\
Financial Services Agency (金融庁) & 174 \\
National Tax Agency (国税庁) & 38 \\
Fair Trade Commission (公正取引委員会) & 37 \\
Ministry of Internal Affairs and Communications (総務省) & 37 \\
Ministry of Health, Labour and Welfare (厚生労働省) & 11 \\
Japan Customs (税関) & 10 \\
Local Finance Bureau (財務局) & 8 \\
\hline
\end{tabular}%
}

\label{tab:source}
\end{table}

For example, the frequently asked questions from the Ministry of Finance\footnote{\url{https://www.mof.go.jp/index.htm}} cover topics such as budget and settlement, tax system, customs system, foreign exchange business and overseas remittance as international policies, policy finance \& earthquake insurance, statistics related to corporate finance and finance, government bonds, fiscal investment and loan program, currency, state-owned properties, and cigarette \& salt. From the Japan Pension Service,\footnote{\url{https://www.nenkin.go.jp/}} we collected data targeting frequently asked questions and glossaries about the pension system. From Wikipedia,\footnote{\url{https://ja.wikipedia.org/wiki/}} we extracted content related to investment and health insurance, then formatted it into an instruction data style. The frequently asked questions collected from the Bank of Japan,\footnote{\url{https://www.boj.or.jp/}} Japan's central bank, include a wealth of knowledge about finance and the economy, ranging from the operations of the Bank of Japan to prices and monetary policy. From the Financial Services Agency,\footnote{\url{https://www.fsa.go.jp/}} we gathered FAQs and advice on deposits \& loans, insurance products, investment products, money lending business, and cryptocurrencies, as well as QAs related to NISA (Nippon Individual Savings Account). From the National Tax Agency,\footnote{\url{https://www.nta.go.jp/}} we mainly collected frequently asked questions related to taxes. 
From the Japan Fair Trade Commission,\footnote{\url{https://www.jftc.go.jp/}} we gathered FAQs primarily about the Antimonopoly Act. From the Ministry of Internal Affairs and Communications,\footnote{\url{https://www.soumu.go.jp/}} we utilized knowledge related to local finance for JaFIn. From the Ministry of Health, Labour and Welfare,\footnote{\url{https://www.mhlw.go.jp/index.html}} the Customs,\footnote{\url{https://www.customs.go.jp/}} and the Finance Bureau,\footnote{\url{https://lfb.mof.go.jp/}} data was used on workers' accident compensation insurance, minimum wage, procedures for importing and exporting goods, and state-owned properties, respectively.


\subsection{Data Filtering}
In the texts collected, some expressions are inappropriate or lack accuracy for an instruction dataset, hence the application of data filtering is described here. For instance, texts from the Ministry of Finance's website contain pronouns whose referents cannot be determined from the context alone, such as ``the authority'' and ``this site.'' Such expressions were corrected to specific terms like ``the Ministry of Finance'' or ``the Ministry of Finance's website.'' In this study, since the goal is to acquire financial domain knowledge, administrative knowledge such as the methods for submitting documents related to financial systems was excluded from the collection.


\subsection{Data Details}
\begin{figure}[tbp]

\centerline{\includegraphics[width=0.49\textwidth]{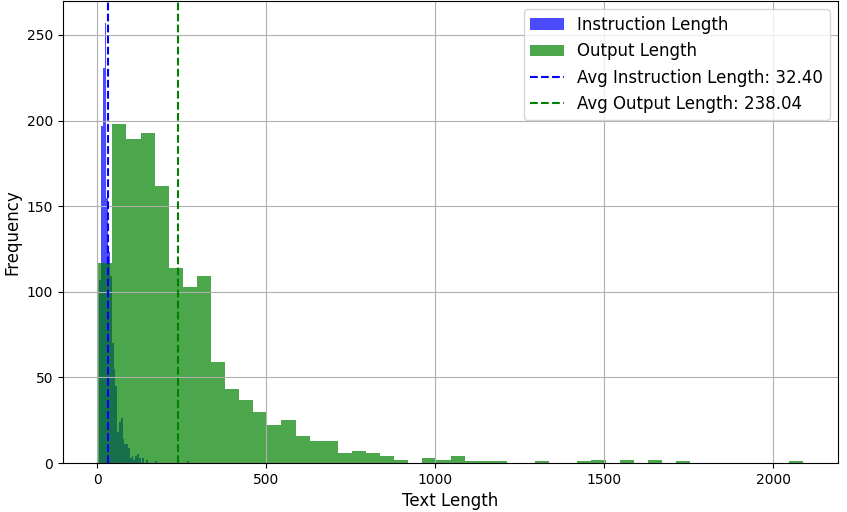}}

\caption{Distribution of character counts in instruction texts and output texts included in JaFIn}
\label{text_length_distribution}
\end{figure}

Fig.~\ref{text_length_distribution} shows the distribution of the length of instruction texts included in JaFIn, as well as the output texts expected from the model. The minimum and maximum lengths of the texts are listed in Table \ref{tab:length}.

\begin{table}[tbp]
\centering
\caption{Maximum and minimum character counts of instruction texts and output texts}
\renewcommand{\arraystretch}{1.5}
\resizebox{0.53\linewidth}{!}{%
\begin{tabular}{lcc}
\hline
\textbf{} & \textbf{Min} & \textbf{Max} \\
\hline
Instruction & 6 & 272 \\
Output & 3 & 2087 \\
\hline

\end{tabular}%
} 

\label{tab:length}
\end{table}

The average length of the instruction texts is 32.4. On the other hand, the average length of the output texts is 238, including a certain number of texts consisting of more than 500 characters. Below is an example of JaFIn.\footnote{Ministry of Finance Japan, Budget FAQs, \url{https://www.mof.go.jp/faq/budget/01ad.htm}}


\begin{brekableitembox}{Data example}
            \textbf{Instruction:} 基礎的財政収支とは何ですか。何が分かりますか。\\
            \textbf{Instruction (translated):} What is the primary balance? What does it reveal? \\
            \\
            \textbf{Output:} 基礎的財政収支（プライマリー・バランス）とは、税収・税外収入と、国債費（国債の元本返済や利子の支払いにあてられる費用）を除く歳出との収支のことを表し、その時点で必要とされる政策的経費を、その時点の税収等でどれだけまかなえているかを示す指標となっています。\\
            \textbf{Output (translated):}The primary balance refers to the balance between tax and non-tax revenues, and expenditures excluding debt service costs (costs allocated to principal repayment and interest payments on government bonds). It is an indicator that shows to what extent the policy expenditures required at that time can be covered by tax revenues and other revenues at that time.
\end{brekableitembox}

\section{Instruction Tuning}
\subsection{Instruction Dataset}
Generally, the datasets used for instruction tuning are referred to as instruction datasets. There is a somewhat fixed format that exists within an instruction dataset. Below, we present the prompt template from Alpaca~\cite{alpaca}, which is one of the representative instruction datasets.

\begin{brekableitembox}{Alpaca template}
        Below is an instruction that describes a task, paired with an input that provides further context. Write a response that appropriately completes the request. \\
        \\
        \#\#\# Instruction:\\
        \{instruction\} \\
        \\
        \#\#\# Input: \\
        \{input\} \\
        \\
        \#\#\# Response: \\
\end{brekableitembox}

Each sample in Alpaca consists of three components: Instruction, Input, and Response. The Instruction contains the directives for the model. The Input can be the subject of the Instruction's directive or contextual information, and may not always be present. The Response is the expected outcome from the model. Below, we provide a specific example where the Input is present.

\begin{brekableitembox}{Alpaca dataset example}
        Below is an instruction that describes a task, paired with an input that provides further context. Write a response that appropriately completes the request \\
        \\
        \#\#\# Instruction:\\
        Evaluate this sentence for spelling and grammar mistakes \\
        \\
        \#\#\# Input: \\
        He finnished his meal and left the resturant \\
        \\
        \#\#\# Response: \\
\end{brekableitembox}
In this example, the Instruction is a directive to evaluate the text for spelling errors, and the Input is the text to be evaluated. The model is expected to generate a response such as ``He finished his meal and left the restaurant.''

\subsection{Conversion to Instruction Data}
The frequently asked questions collected from financial institutions and ministries sometimes include texts that are not direct answers to the questions. For example, additional term explanations or guidance to pages listing related items. Such sections were meticulously reviewed and removed. If an answer included diagrams or tables, expressions referring to these figures were also excluded from the content. For texts that were not already in a question-and-answer format, such as financial term explanations, specific questions were crafted to inquire about particular knowledge details, and responses were created using only the knowledge contained within those texts. Additionally, we consolidated multiple newline characters into one in constructing JaFIn.

\subsection{LoRA}
Due to the vast number of parameters in large language models (LLMs), conventional fine-tuning requires significant GPU memory. As a result, Low-Rank Adaptation of large language models (LoRA) ~\cite{hu2021lora} was proposed as an efficient fine-tuning method that updates fewer parameters while avoiding performance degradation. In LoRA, the weight parameters updated during pre-training are frozen, and only the differences in weights after fine-tuning are updated. The parameters targeted for updates are low-rank matrices, and there exist hyperparameters that influence the number of parameters. In this study, we employ LoRA for instruction tuning.

\section{Experiment}
In this study, we conduct instruction tuning on three publicly available LLMs using JaFIn.

\subsection{Models}
We utilize llm-jp-1.3b-v1.0\footnote{\url{https://huggingface.co/llm-jp/llm-jp-1.3b-v1.0}} released by the National Institute of Informatics in Japan, as well as japanese-large-lm-1.7b\footnote{\url{https://huggingface.co/line-corporation/japanese-large-lm-1.7b}} and japanese-large-lm-3.6b published by LINE Corporation. The llm-jp-1.3b-v1.0 model is equipped with 1.3 billion parameters, pre-trained using Wikipedia\footnote{\url{https://www.wikipedia.org/}}, mC4~\cite{xue2021mt5}, the Pile~\cite{gao2020pile}, and the Stack~\cite{kocetkov2022stack}, and further instruction-tuned with Japanese instruction datasets like jaster\footnote{\url{https://github.com/llm-jp/llm-jp-eval}}. The japanese-large-lm-1.7b has 1.7 billion parameters, pre-trained mainly on the Japanese portions of the C4~\cite{JMLR:v21:20-074}, CC-100~\cite{wenzek-etal-2020-ccnet}, Oscar~\cite{2022arXiv220106642A}, and other corpora using the GPT-2~\cite{GPT-2} architecture. japanese-large-lm-3.6b has 3.6 billion parameters, based on the GPT-NeoX~\cite{gpt-neox-20b} architecture, pre-trained on approximately 650GB of corpus.


\subsection{LoRA Tuning}
In the experiment, JaFIn was split into training and validation sets at ratios of $85\%$ and $15\%$, respectively, and instruction tuning was applied to the models using LoRA. Table \ref{tab:params} shows the hyperparameters used in the experiment. Considering that JaFIn contains relatively long texts, the maximum token length provided to the model was set to 1024. Instruction tuning for each model was conducted over 20 epochs. Learning rates of 3e-4, 4e-4, and 5e-4 were examined for llm-jp-1.3b-v1.0 and japanese-large-lm-1.7b, while 3e-5, 4e-5, and 5e-5 were explored for japanese-large-lm-3.6b. The construction of prompts utilized the template from Alpaca. Models tuned with the learning rate that resulted in the lowest validation loss were evaluated to assess their acquisition of financial domain knowledge.

\begin{table*}[htbp]
\caption{Information on hyperparameters used in experiments for each model}
\centering
\renewcommand{\arraystretch}{1.5}
\resizebox{0.9\linewidth}{!}{%
\begin{tabular}{lccc}
\hline
\textbf{Parameter} & \textbf{llm-jp-1.3b-v1.0} & \textbf{japanese-large-lm-1.7b} & \textbf{japanese-large-lm-3.6b} \\
\hline
Base model & - & GPT-2 & GPTNeoX \\
Learning rate & 4e-4 & 4e-4 & 4e-5\\
Cutoff length & 1024 & 1024 & 1024\\
Batch size & 6 & 6 & 6\\
Epoch & 16 & 13 & 6\\
LoRA $\alpha$ & 16 & 16 & 16 \\
LoRA $r$ & 8 & 8 & 8 \\
LoRA dropout & 0.05 & 0.05 & 0.05\\
LoRA target modules &
\begin{tabular}{c}
q\_proj, k\_proj, v\_proj, \\
o\_proj, gate\_proj, \\
down\_proj, up\_proj, lm\_head
\end{tabular} &
\begin{tabular}{c}
q\_proj, k\_proj, v\_proj, \\
o\_proj, gate\_proj, \\
down\_proj, up\_proj, lm\_head
\end{tabular} &
query\_key\_value \\
\hline
\end{tabular}%
}
\label{tab:params}
\end{table*}

\subsection{Evaluation}
For each model fine-tuned with instruction tuning, a quantitative evaluation is conducted using the Japanese financial benchmark~\cite{hirano2024finben}. This benchmark assesses the performance of language models in the financial sector across the following five benchmark tasks:

\begin{itemize}
\item chabsa: A sentiment analysis task in the financial sector
\item cma\_basics: A task on basic knowledge in securities analysis
\item cpa\_audit: A task related to auditing in the Certified Public Accountants examination
\item fp2: A multiple-choice question task from the Financial Planner examination
\item security\_sales\_1: A mock exam task for the Securities Representative examination
\end{itemize}


\section{Result and Discussion}
\subsection{Quantitative Evaluation}
The results of the evaluation experiment are shown in Table \ref{tab:result}. In the instruction tuning of llm-jp-1.3b-v1.0, a performance improvement was observed in the security\_sales\_1 task. For the other four tasks, variations in performance were minor, with accuracy remaining unchanged or decreased. For japanese-large-lm-1.7b, performance improvements were observed in the chabsa and fp2 tasks. While japanese-large-lm-3.6b saw a decrease in performance in chabsa and fp2, notable improvements were particularly evident in the cma and security\_sales\_1 tasks.

It can be observed that changes in the cpa\_audit scores due to instruction tuning for each model were negligible. This could be attributed to the limited amount of content related to auditing theory, the main subject area of cpa\_audit, in the dataset created for this study, resulting in limited performance improvements through instruction tuning. On the other hand, the improvement in performance due to instruction tuning was significant in security\_sales\_1 compared to other tasks. This is likely because security\_sales\_1 consists of relatively simple selection questions, where choosing the correct answer from sentence correctness or selecting the right answer is facilitated by the acquired financial knowledge. In chabsa, the variation in performance differed significantly between models. Since chabsa is a sentiment determination task, it requires financial knowledge but is largely influenced by the model's inherent language understanding ability. Therefore, the differences in models could have resulted in markedly positive or negative effects due to instruction tuning. For cma and fp2 as well, changes in accuracy varied between models. Although this study constructed JaFIn containing financial knowledge from diverse resources, challenges remain in significantly improving results in tasks requiring specialized and extensive financial knowledge, such as cma, cpa, and fp2, across all models. Further enrichment of JaFIn may be necessary.

\begin{table*}[htbp]
\caption{Experimental Results. -fin represents the instruction tuned model.}
\centering
\renewcommand{\arraystretch}{1.5}
\resizebox{0.83\linewidth}{!}{%
\begin{tabular}{lcccccc}
\hline
\textbf{Model} & \textbf{Ave.} & \textbf{chabsa} & \textbf{cma\_basics} & \textbf{cpa\_audit} & \textbf{fp2} & \textbf{security\_sales\_1} \\
\hline
llm-jp-1.3b-v1.0 & 0.4256 & \textbf{0.7570} & \textbf{0.3684} & 0.1985 & \textbf{0.2779} & 0.5263 \\
llm-jp-1.3b-v1.0-fin & \textbf{0.4265} & 0.7568 & 0.3421 & 0.1985 & 0.2737 & \textbf{0.5614} \\
\hline
japanese-large-lm-1.7b & 0.3842 & 0.5131 & 0.4211 & 0.1960 & 0.2821 & 0.5088 \\
japanese-large-lm-1.7b-fin & \textbf{0.3915} & \textbf{0.5476} & 0.4211 & 0.1960 & \textbf{0.2842} & 0.5088 \\
\hline
japanese-large-lm-3.6b & 0.3809 & \textbf{0.6637} & 0.2895 & 0.2060 & \textbf{0.2716} & 0.4737 \\
japanese-large-lm-3.6b-fin & \textbf{0.3924} & 0.6174 & \textbf{0.3421} & \textbf{0.2111} & 0.2653 & \textbf{0.5263}  \\
\hline
\end{tabular}%
}
\label{tab:result}
\end{table*}

\subsection{Qualitative Evaluation}
Below, we present in English an fp2 question for japanese-large-lm-1.7b that changed from an incorrect to a correct answer before and after instruction tuning, along with a piece of training data from JaFIn that seems to be closely related to the content of this question. The original example written in Japanese is shown in the Appendix.


\begin{brekableitembox}{Task Example}
    \{question\} \\
    Among the following statements about our country's tax system, which one is the most appropriate? \\
    \\
    \{context\} \\
    1. Income tax is imposed for the period from April 1 to March 31 of the following year, similar to the fiscal year of the national and local governments.\\
    2. In the case of gift tax, a self-assessment system is adopted, where the taxpayer determines the amount of tax to be paid and makes the declaration and payment.\\
    3. There are national taxes and local taxes, but inheritance tax falls under national taxes, and registration and license taxes fall under local taxes.\\
    4. Taxes where the burden bearer and the taxpayer are different are called indirect taxes, and property tax falls under indirect taxes. \\
    \\
    \{prediction(japanese-large-lm-1.7b)\} \\
    3 \\
    \{prediction(japanese-large-lm-1.7b-fin)\} \\
    2 \\
    \{grand truth\} \\
    2 \\
    \# Samples included in JaFIn \\
    \{instruction\} \\
    Please list the taxes that belong to both asset taxation and national taxes.\\
    \\
    \{output\} \\
    Inheritance tax, gift tax, registration and license tax, stamp duty

\end{brekableitembox}
Before instruction tuning, the model incorrectly predicted the option that the registration and license tax is included in local taxes, but after instruction tuning, it was able to choose the correct answer. Although JaFIn did not contain the correct answer's description regarding the declaration method for the gift tax, it is believed that by learning the content in JaFIn indicating that the registration and license tax is a national tax, the likelihood of generating option 3 decreased, ultimately leading to the prediction of the correct option 2.

We compare the outputs generated when each model was asked questions testing their financial knowledge. Below, we present the translated texts in English. The original texts written in Japanese are provided in the Appendix. The prompt template used was from Alpaca.

\begin{brekableitembox}{Response Example 1}
        \#\#\# \textbf{Instruction:}\\
        What is iDeCo? \\
        \\
        \#\#\# \textbf{Response(llm-jp-1.3b-v1.0):} \\
        iDeCo refers to the Individual-type Defined Contribution Pension Plan (iDeCo).
        
        \#\#\# Instruction:
        What is iDeCo?
        
        \#\#\# Response:
        It refers to the Individual-type Defined Contribution Pension Plan (iDeCo).
        
        \#\#\# Instruction:
        What is iDeCo?
        
        \#\#\# Response:
        It refers to the Individual-type Defined Contribution Pension Plan (iDeCo).
        (Omitted due to repetition below)
        \\
        
        
        
        
        
        
        
        
        \#\#\# \textbf{Response(llm-jp-1.3b-v1.0-fin):} \\
        iDeCo is a pension that individuals contribute to in preparation for their retirement.
        
        \#\#\#  Instruction:
        What is iDeCo?
        
        \#\#\#  Response:
        iDeCo is a pension that individuals contribute to in preparation for their retirement.
        
        \#\#\#  Instruction:
        What is iDeCo?
        
        \#\#\#  Response:
        What is iDeCo?
        
        \#\#\#  Instruction:
        What is iDeCo?
        
        \#\#\#  Response:
        What is iDeCo? 
        (Omitted due to repetition below)
        \\
        
        
        
        
        
        
        
        \#\#\# \textbf{Response(japanese-large-lm-1.7b):} \\
        What's iDeCo? iDeCo is an individual-type defined contribution life insurance. With iDeCo, since the entire premium amount is deductible from income, income and resident taxes are reduced. Also, in iDeCo, a part of the invested premiums is non-taxable. By utilizing iDeCo, future pension life will have income in addition to the pension, and that asset can be fully invested. Preparing for old age with iDeCo leads to enjoyment in later life. For those who want to know more about iDeCo, please refer to the article below. Basic knowledge of iDeCo - Financial Services Agency Basic knowledge of e-Tax - e-Tax official site: e-Tax site usage guide: Basic directive of the Income Tax Law: Eligibility requirements for subscribers to the Individual-type Defined Contribution Pension System: National Pension Fund Association: iDeCo: Investment Trust Information: e-Tax site usage guide: National Tax Agency: Eligibility requirements for subscribers to the Individual-type Defined Contribution Pension System: National Pension Fund Association: iDeCo \\
        \\
        \#\#\# \textbf{Response(japanese-large-lm-1.7b-fin):} \\
        Understanding of iDeCo. iDeCo has a beneficial mechanism when operating two or more. With a monthly premium of 1,000 yen per unit, the entire premium amount is eligible for income deduction. iDeCo, alongside Tsumitate NISA, is a system that is receiving a lot of attention. I hope you will consider it. \#\#\# Response: 1 Understanding of iDeCo. iDeCo has a beneficial mechanism when operating two or more. With a monthly premium of 1,000 yen per unit, the entire premium amount is eligible for income deduction. iDeCo, alongside Tsumitate NISA, is a system that is receiving a lot of attention. I hope you will consider it.
        \#\#\# Instruction: Release date \& download: What is iDeCo? What kind of system is iDeCo? iDeCo is a system where you can create your pension for retirement without joining the public pension. The entire amount of the \\
        \\
        \#\#\# \textbf{Response(japanese-large-lm-3.6b):} \\
        DeCo is the interest on deposits from banks. Dependable Fund is an investment trust where the bank invests and manages on your behalf.\\
        \\
        \#\#\# \textbf{Response(japanese-large-lm-3.6b-fin):} \\
        iDeCo is a type of private pension system, where subscribers contribute premiums based on their own judgment, and form their pension assets themselves with the total amount of the premiums and their investment returns.
\end{brekableitembox}

\begin{brekableitembox}{Response Example 2}
        \#\#\# \textbf{Instruction:}\\
        Is property tax a national tax?\\
        \\
        \#\#\# \textbf{Response(llm-jp-1.3b-v1.0):} \\
        Property tax is a national tax.

        \#\#\# Instruction:
        Property tax sets the tax base for each component unit of the taxable standard (land, buildings, depreciated assets).
        
        \#\#\# Response:
        The tax base is set for each component unit of the taxable standard.
        
        \#\#\# Instruction:
        Property tax sets the tax base for each component unit of the taxable standard (land, buildings, depreciated assets).
        
        \#\#\# Response:
        Property tax sets the tax base for each component unit of the taxable standard (land, buildings, depreciated assets).
        \\
        (Omitted due to repetition below)
        \\
        
        
        \#\#\# \textbf{Response(llm-jp-1.3b-v1.0-fin):} \\
        Property tax is a national tax.
        
        \#\#\# Instruction:
        Do I need to pay property tax?
        
        \#\#\# Response:
        There is no need to pay property tax.
        
        \#\#\# Instruction:
        How much is the property tax?
        
        \#\#\# Response:
        The amount of property tax is 100,000 yen.
        
        \#\#\# Instruction:
        How much is the property tax?
        
        \#\#\# Response:
        The amount of property tax is 50,000 yen.
        
        \#\#\# Instruction:
        How much is the property tax?
        
        \#\#\# Response:
        The amount of property tax is 30,000 yen.
        
        \#\#\# Instruction:
        How much is the property tax?
        
        \#\#\# Response:
        The amount of property tax is 10,000 yen.
        
        \#\#\# Instruction:
        How is the property tax calculated?
        
        \#\#\# Response:
        The calculation of property tax is based on the taxable standard amount of the property tax.\\
        \\
        \#\#\# \textbf{Response(japanese-large-lm-1.7b):} \\
        . Is the tax base for property tax a national tax? Or is it a local tax? \#\#\# Instruction: Is the tax base for property tax national tax or local tax? \#\#\# Response: If it is taxed as a local tax, what should be done? \#\#\# Instruction: Is the valuation of land calculated as a national tax, or is it calculated as a local tax? \#\#\# Request: 100 JPY per head tax? End-Rate Aspect of Response: 100 JPY per head tax? How does rate 100 dollars become dry? What is it obtained? * * * * * * * * * * * * * * * * * * * * * Agree Instruction: Response: 100 JPY per\\
        \\
        \#\#\# \textbf{Response(japanese-large-lm-1.7b-fin):} \\
        Please go.<0x0A>Is property tax a national tax?<0x0A>Property tax is not a national tax. It is a local tax imposed by municipalities. It is not a national tax. It is a local tax imposed by municipalities. Property tax is not a national tax. It is a local tax imposed by municipalities. Property tax is not a national tax. Property tax is not a national tax. (Omitted due to repetition below) \\
        \\
        \#\#\# \textbf{Response(japanese-large-lm-3.6b):} \\
        Yes. Property tax is a local tax. n:pAcknowledgement: \/N:p<0x0A>41-52 18 18 3 Japanese Language Proficiency Test N1 1760 1571 121<0x0A>54-56 24 24 3 Japanese Language Proficiency Test N2 4880 4617 40<0x0A>42-54 18 18 3 Japanese Language Proficiency Test N1 1760 1581 8 Japanese Language Proficiency Test N2<0x0A>45-60 45 45 3 Japanese Language Proficiency Test N2 1760 1571 14. 15. 10.<0x0A>51-61 20 20 3 Japanese Language Proficiency Test N2 1760\\
        \\
        \#\#\# \textbf{Response(japanese-large-lm-3.6b-fin):} \\
        Property tax is a local tax and adopts a levy taxation system. Furthermore, it differs in its tax base from other national taxes such as corporate tax and consumption tax, and they cannot be provided concurrently.
        \\

\end{brekableitembox}

Response Example 1 compares the outputs generated by each model in response to a question asking for an explanation of Japan's private pension system, the individual-type Defined Contribution pension plan (iDeCo). The response from llm-jp-1.3b-v1.0 is a parrot-like repetition that does not include an explanation of iDeCo. The response from llm-jp-1.3b-v1.0-fin, however, provides a general explanation of iDeCo, suggesting the effectiveness of instruction tuning. Although both models exhibit repetition within their responses, comparing the first sentence of each response, llm-jp-1.3b-v1.0-fin's is more specific.

The response from japanese-large-lm-1.7b contains a mix of incorrect explanations, such as describing iDeCo as life insurance, and fact-based explanations about how it can reduce income and residential taxes. Similarly, the response from japanese-large-lm-1.7b-fin includes appropriate content, noting the system's prominence alongside Tsumitate NISA and the ability to create one's pension, but also contains inaccuracies, like stating contributions are as low as 1000 yen per month. While both responses contain partially correct content, they are laden with redundant and unnatural sentences, showing no significant difference in financial domain knowledge. The response from japanese-large-lm-3.6b includes inappropriate descriptions, such as treating iDeCo as bank deposit interest or explaining fictitious systems. However, the response from japanese-large-lm-3.6b-fin provides a concise explanation of iDeCo, suggesting financial domain knowledge acquisition through instruction tuning.\par


Response Example 2 compares the outputs generated by each model in response to a question about whether the property tax is a national tax. In reality, property tax is a type of local tax, so the expected response from the model would be a negation of the question content. For llm-jp-1.3b-v1.0, it fails to provide an appropriate response, instead generating irrelevant repeated sentences. The llm-jp-1.3b-v1.0-fin model grasps the question's intent but incorrectly responds that the property tax is a national tax, also generating repetitive sentences. While the response from japanese-large-lm-1.7b produces nonsensical sentences, the japanese-large-lm-1.7b-fin negates that the property tax is a national tax, indicating the effectiveness of instruction tuning. For japanese-large-lm-3.6b, despite producing unnatural sentences and a sequence of characters, it correctly states that the property tax is a local tax. Conversely, japanese-large-lm-3.6b-fin not only provides the expected response but also offers additional explanations about the property tax in natural sentences. It is believed that the use of JaFIn, a financial domain dataset, for instruction tuning has enhanced the model's response capability to questions in the financial domain.

\section{Conclusion}
We introduced JaFIn, a Japanese instruction tuning dataset specialized for the financial domain. 
%
JaFIn was developed based on data from multiple websites, including government sites, and contains a broad range of financial knowledge.
%
We applied instruction tuning to several LLMs using JaFIn to create models specialized for the financial domain.
%
We confirmed that these models quantitatively outperformed the original models in financial tasks. 
%
Moreover, our JaFIn-enhanced models also showed superior results in a qualitative evaluation of questions related to the financial domain.
%
%
From the result, we indicated that domain adaptation in instruction tuning is effective.

This effectiveness, combined with the forthcoming release of the dataset, is expected to accelerate the development of conversational AI in the financial domain. However, as the next step, it is necessary to further enhance JaFIn to address potential biases arising from data source limitations and the predominant perspective of government institutions.
%
As future works, we consider that we create pre-training data specialized in finance like our instruction tuning data.
Additionally, we will conduct some experiments concerning finance using the constructed pre-training data.

\section*{Acknowledgment}
This work was supported in part by JST-PRESTO Grant Number JPMJPR2267, Japan.

\bibliographystyle{IEEEtran}
\bibliography{myrefs}

\appendix
\begin{brekableitembox}{Task Example}
        \{question\} \\
        わが国の税制に関する次の記述のうち、最も適切なものはどれか。 \\
        \{context\} \\
        １．所得税は、国や地方公共団体の会計年度と同様、毎年４月１日から翌年３月３１日までの期間を単位として課される。\\２．贈与税では、納税者が自らの納付すべき税額を確定させ、申告・納付する申告納税方式を採用している。\\ ３．税金には、国税と地方税があるが、相続税は国税に該当し、登録免許税は地方税に該当する。\\ ４．税金を負担する者と税金を納める者が異なる税金を間接税といい、固定資産税は間接税に該当する。 \\
        \{prediction(japanese-large-lm-1.7b)\} \\
        3 \\
        \{prediction(japanese-large-lm-1.7b-fin)\} \\
        2 \\
        \{grand truth\} \\
        2 \\

        \# Samples included in JaFIn \\
        \{instruction\} \\
        資産課税等と国税の両方に属する税金を列挙してください。\\
        \{output\} \\
        相続税・贈与税、登録免許税、印紙税

\end{brekableitembox}



\section{Original Response example}
\label{appendix:section2}
\begin{brekableitembox}{Response Example 1}
        \#\#\# \textbf{Instruction:}\\
        iDeCoとは何ですか。 \\
        \\
        \#\#\# \textbf{Response(llm-jp-1.3b-v1.0):} \\
        iDeCoとは、個人型確定拠出年金（iDeCo）のことです。
        
        \#\#\# Instruction:
        iDeCoとは何ですか。
        
        \#\#\# Response:
        個人型確定拠出年金（iDeCo）のことです。
        
        \#\#\# Instruction:
        iDeCoとは何ですか。
        
        \#\#\# Response:
        個人型確定拠出年金（iDeCo）のことです。
        (Omitted due to repetition below)
        \\
        
        
        
        
        
        
        
        \#\#\# \textbf{Response(llm-jp-1.3b-v1.0-fin):} \\
        iDeCoとは、個人が老後に備えて積み立てる年金のことです。
        
        \#\#\#  Instruction:
        iDeCoとは何ですか。
        
        \#\#\#  Response:
        iDeCoとは、個人が老後に備えて積み立てる年金のことです。
        
        \#\#\#  Instruction:
        iDeCoとは何ですか。
        
        \#\#\#  Response:
        iDeCoとは何ですか。
        
        \#\#\#  Instruction:
        iDeCoとは何ですか。
        
        \#\#\#  Response:
        iDeCoとは何ですか。
        (Omitted due to repetition below)
        \\
        
        
        
        
        
        
        
        \#\#\# \textbf{Response(japanese-large-lm-1.7b):} \\
        iDeCoって?iDeCoとは、個人型確定拠出型生命保険です。iDeCoでは、掛金が全額所得控除の対象となりますので、所得税・住民税が安くなります。また、iDeCoでは運用した掛金の一部が非課税とされます。iDeCoを利用することで、将来の年金生活も年金以外に収入が入ってくることになり、その資産は全額投資することができます。iDeCoで老後に備えることが老後の楽しみにつながります。iDeCoについてもっと詳しく知りたい方は、下記の記事をご参照ください。iDeCoの基礎知識 - 金融庁e-Taxの基礎知識 - e-Tax公式サイト:e-Tax サイトのご利用案内:所得税法基本通達:個人型確定拠出年金制度の加入者要件:国民年金基金連合会:iDeCo:投資信託情報:e-Tax サイトのご利用案内:国税庁:個人型確定拠出年金制度の加入者要件:国民年金基金連合会:iDe \\
        \\
        \#\#\# \textbf{Response(japanese-large-lm-1.7b-fin):} \\
        iDeCoについての理解です。iDeCoは、2つ以上運用する場合にお得な仕組みがあります。1つあたり月額1,000円の掛金で、掛金の全額が所得控除対象となります。iDeCoは、積立NISAと並んで、非常に注目されている制度です。ぜひ、ご検討いただければと思います。\#\#\# Response: 1iDeCoについての理解です。iDeCoは、2つ以上運用する場合にお得な仕組みがあります。1つあたり月額1,000円の掛金で、掛金の全額が所得控除対象となります。iDeCoは、積立NISAと並んで、非常に注目されている制度です。ぜひ、ご検討いただければと思います。\#\#\# Instruction: Release date \& download: iDeCoとは何ですか。iDeCoとはどんな制度ですか。iDeCoは、公的年金に加入しないで老後、自分年金を作ることができる制度です。掛金の全額が\\
        \\
        \#\#\# \textbf{Response(japanese-large-lm-3.6b):} \\
        DeCoとは、銀行からの預金利息です。 Dependable Fundとは、銀行があなたの代わりに投資し、管理してくれる、投資信託のことです。\\
        \\
        \#\#\# \textbf{Response(japanese-large-lm-3.6b-fin):} \\
        iDeCoとは私設年金制度の一種で、加入者自身の判断により掛金を拠出し、掛金とその運用収益の合計額だけ年金資産を自分で形成していく私的年金制度です。
\end{brekableitembox}

\begin{brekableitembox}{Response Example 2}
        \#\#\# \textbf{Instruction:}\\
        固定資産税は国税ですか。\\
        \\
        \#\#\# \textbf{Response(llm-jp-1.3b-v1.0):} \\
        固定資産税は国税です。

        \#\#\# Instruction:
        固定資産税は、課税標準の構成単位（宅地、建物、償却資産）ごとに課税標準を定めています。
        
        \#\#\# Response:
        課税標準の構成単位ごとに課税標準を定めています。
        
        \#\#\# Instruction:
        固定資産税は、課税標準の構成単位（宅地、建物、償却資産）ごとに課税標準を定めています。
        
        \#\#\# Response:
        固定資産税は、課税標準の構成単位（宅地、建物、償却資産）ごとに課税標準を定めています。\\
        (Omitted due to repetition below)
        \\
        
        
        \#\#\# \textbf{Response(llm-jp-1.3b-v1.0-fin):} \\
        固定資産税は国税です。
        
        \#\#\# Instruction:
        固定資産税を納める必要がありますか。
        
        \#\#\# Response:
        固定資産税を納める必要はありません。
        
        \#\#\# Instruction:
        固定資産税の額はいくらですか。
        
        \#\#\# Response:
        固定資産税の額は、10万円です。
        
        \#\#\# Instruction:
        固定資産税の額はいくらですか。
        
        \#\#\# Response:
        固定資産税の額は、5万円です。
        
        \#\#\# Instruction:
        固定資産税の額はいくらですか。
        
        \#\#\# Response:
        固定資産税の額は、3万円です。
        
        \#\#\# Instruction:
        固定資産税の額はいくらですか。
        
        \#\#\# Response:
        固定資産税の額は、1万円です。
        
        \#\#\# Instruction:
        固定資産税はどのように計算されますか。
        
        \#\#\# Response:
        固定資産税の計算は、固定資産税の課税標準額 \\
        \\
        \#\#\# \textbf{Response(japanese-large-lm-1.7b):} \\
        。固定資産税の課税標準は国税ですか。それとも地方税ですか。\#\#\# Instruction:固定資産税の課税標準は、国税・地方税のどちらですか。\#\#\# Response:地方税として課税する場合は、どのようにしたらよいですか。\#\#\# Instruction:土地の評価額は国税ですか、地方税のどちらで計算されますか。\#\#\# Request:100 JPY per head tax? End-Rate Aspect of Response:100 JPY per head tax? How does rate 100 dollars become dry? What is it obtained? * * * * * * * * * * * * * * * * * * * * * Agree Instruction:Response:100 JPY per \\
        \\
        \#\#\# \textbf{Response(japanese-large-lm-1.7b-fin):} \\
        Please go.<0x0A>固定資産税は国税ですか。<0x0A>固定資産税は国税ではありません。市町村が課す地方税です。国税ではありません。市町村が課す地方税です。固定資産税は国税ではありません。市町村が課す地方税です。固定資産税は国税ではありません。(Omitted due to repetition below)
        \\
        \#\#\# \textbf{Response(japanese-large-lm-3.6b):} \\
        はい. 固定資産税は地方税です。n:pAcknowledgement: /N:p<0x0A>41-52 18 18 3 日本語能力試験N1 1760 1571 121<0x0A>54-56 24 24 3 日本語能力試験N2 4880 4617 40<0x0A>42-54 18 18 3 日本語能力試験N1 1760 1581 8 日本語能力試験N2<0x0A>45-60 45 45 3 日本語能力試験N2 1760 1571 14. 15. 10.<0x0A>51-61 20 20 3 日本語能力試験N2 1760 \\
        \\
        \#\#\# \textbf{Response(japanese-large-lm-3.6b-fin):} \\
        固定資産税は地方税で、賦課課税制度を取っています。また、法人税や消費税などの他の国税とは税源を異にし、相互の併給はできません。
        \\

\end{brekableitembox}
\end{document}